\documentclass[letterpaper, 10 pt, conference]{ieeeconf}  

\IEEEoverridecommandlockouts                              

\usepackage{times}
\usepackage{epsfig}
\usepackage{graphicx}
\usepackage{amsmath}
\usepackage{amssymb}
\usepackage{multirow}
\usepackage{centernot}
\usepackage[table,xcdraw]{xcolor}
\usepackage{subcaption}

\title{\LARGE \bf
Extending 3D body pose estimation for robotic-assistive therapies of autistic children*
}

\author{Laura Santos$^{1,2}$, Bernardo Carvalho$^{1}$, Catarina Barata$^{1}$, José Santos-Victor$^{1}$
\thanks{*This work was supported by the FCT Portuguese Foundation of Science and Technology (projects SFRH/BD/145040/2019, CEECIND/00326/2017,  UID/50009/2020 LARSyS, C645008882-00000055 Center for Responsible AI, and the ELLIS Lisbon Unit)}
\thanks{$^{1}$ Institute for Systems and Robotics, LARSyS, Instituto Superior T\'ecnico, Universidade de Lisboa, Portugal.
        {\{laura.d.santos, bernardosmfcarvalho, ana.c.fidalgo.barata\}@tecnico.ulisboa, jasv@isr.tecnico.ulisboa}}%
\thanks{$^{2}$NEARLab, Department of Electronics, Information and Bioengineering, Politecnico di Milano, Milan, Italy.}%
}
\begin{document}

\maketitle
\thispagestyle{empty}
\pagestyle{empty}

\begin{abstract}

Robotic-assistive therapy has demonstrated very encouraging results for children with Autism. Accurate estimation of the child's pose is essential both for human-robot interaction and for therapy assessment purposes. Non-intrusive methods are the sole viable option since these children are sensitive to touch. 
While depth cameras have been used extensively, existing methods face two major limitations: (i) they are usually trained with adult-only data and do not correctly estimate a child's pose, and (ii) they fail in scenarios with a high number of occlusions. 
Therefore, our goal was to develop a 3D pose estimator for children, by adapting an existing state-of-the-art 3D body modelling method and incorporating a linear regression  model to fine-tune one of its inputs, thereby correcting the pose of children's 3D meshes. 
In controlled settings, our method has an error below $0.3m$, which is considered acceptable for this kind of application and lower than current state-of-the-art methods. In real-world settings, the proposed model performs similarly to a Kinect depth camera and manages to successfully estimate the 3D body poses in a much higher number of frames.
\end{abstract}

\section{Motivation}

Robotic therapy has been used in Autism Spectrum Disorder (ASD) treatment since children with this disorder have shown a high interest in robots, demonstrating similar behaviours to those healthy children have towards adults \cite{multi-robot_therapy}. Given the heterogeneity of this disorder, different types of therapy have been developed to train motor, social and cognitive skills such as imitation, emotion recognition, and attention. In these therapies, the robot and the child are always present and the therapist can have an active role in participating in the activities suggested by the robot or just be a supporter of the child. In this context, pose reconstruction of the participants is important for two reasons: on one hand, it allows the evaluation of performance and understanding of how the subjects are evolving during the sessions; on the other hand, the knowledge of the positions is required for the interaction of the robot with the participants, for example during imitation games \cite{robot_imitation}. For this type of interaction, depth cameras are used since they are not intrusive to children, a requirement for children with this disorder.

\begin{figure}[!h]
	\centering
	\includegraphics[width=0.3\textwidth]{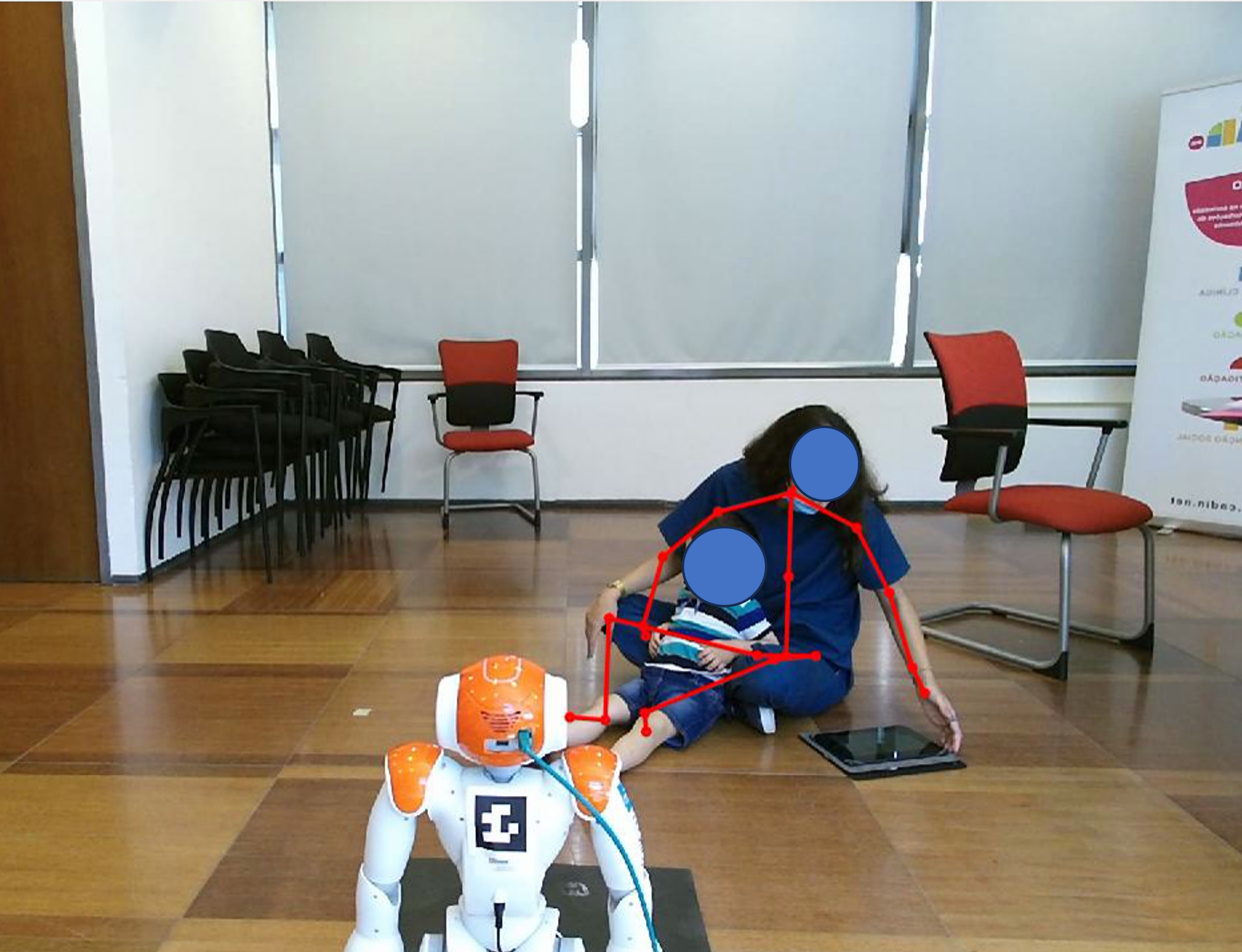}
	\caption{Example of challenging position during therapy captured by a depth camera. The represented skeleton is a hybrid of the upper body of the therapist and the lower body of the child.} 
	\label{problem}
\end{figure}

Typically the environment where the therapeutic sessions take place is unconstrained, meaning that children can move freely, creating occlusions either during the interaction with the therapist or by the particular execution of certain movements (Figure \ref{problem}). This problem can become even more complex, as the same therapeutic protocol may be applied to children of a big range of ages (from 2 to 10 years old) and by therapists whose approaches to therapy can differ depending on the speciality (\textit{e.g.}, speech therapy, physiotherapy), creating a range of occlusions. Thus, a depth camera often fails to estimate the skeletons of the subjects in the session or estimates incorrect ones. An alternative can be offline pose reconstruction. In literature, several models are available, but they are trained principally in adults, showing several errors when estimating children. Moreover, few models explore the interactions between children and adults. 

Thus, the main goal of this work is the adaptation of a 3D pose estimator to children to be used in a therapeutic scenario, first in an offline manner for the recovery of poses lost during a session, and secondly in an online manner to allow the interaction between a robot and a child.

We start by reviewing the main models used in the literature, then we present our improved version and the tests we made both in constrained and unconstrained scenarios.

\section{State of the art}
\label{sec:stateofart}
During robotic-assistive therapy, participants' skills have to be assessed to understand their impact. The most common measures evaluated and integrated into the robot control regard patient engagement and motor skills\cite{Marinoiu_2018_CVPR, Rehg_2013,  robot_imitation}.
Different types of measures are available and used for motor skills evaluation. However, children with ASD are sensitive to touch, so non-intrusive solutions are preferred \cite{robot_imitation} to more intrusive ones. Moreover, these children move considerably during therapies, which is another restriction to intrusive solutions. Therefore, LIDAR and depth cameras are preferred, such as Microsoft Kinect known as Kinect \cite{multi-robot_therapy, Marinoiu_2018_CVPR}.

\begin{figure}[!h]
	\centering
	\includegraphics[width=0.35\textwidth]{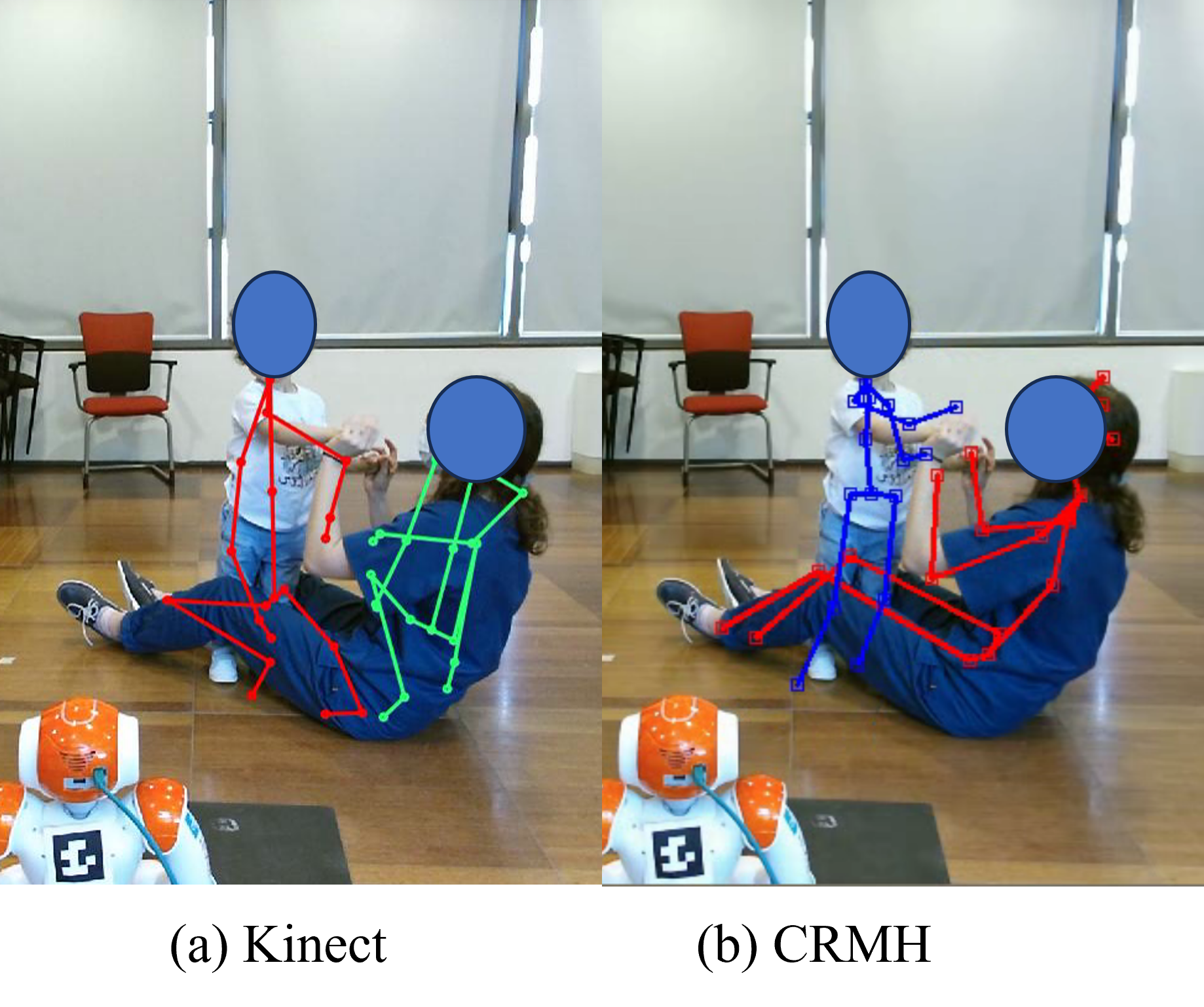}
	\caption{Kinect and CRMH outputs for an image with body occlusions. While the Kinect skeletons for the therapist and child are affected by the severe number of self and interpersonal occlusions, the CRMH model is able to effectively reconstruct the two skeletons.} 
	\label{bodyocclusions}
\end{figure}

Kinect has its own algorithm to identify people and their respective 3D skeletons, but this system has problems when dealing with occlusions (either self or interpersonal ones) \cite{Nguyen2022}. Recently, several works proposed approaches to estimate the 3D skeleton or 3D shape in a multiperson scenario, given 2D images. An example is in Mustafa et al. \cite{Mustafa_2021_CVPR}. The authors focus on the 3D reconstruction of several people in a given scenario, for example, a street full of people walking. They use a multitask network to learn an implicit 3D reconstruction of the scenario and identify the spatial location and orientation of each person. It can deal with complex poses and partial occlusions, but there are no demonstrated situations where people were seated or had some interaction between them. Similarly, Jiang et al. \cite{Jiang_2020_CVPR} propose the Coherent Reconstruction of multiple humans (CRMH) method that considers the coherent recreation of the scene and the possible interactions between people simultaneously, using an end-to-end framework for 3D pose and shape estimation of all the individuals in an image. In summary, the CRMH method receives as input the image and a focal length parameter, and returns as output the 3D mesh of the people in the image. Its model was trained with challenging positions, considering occlusions between people, and being able to deal with them (Figure \ref{bodyocclusions}). 
 

The CRMH system has two blocks. In the detection block, the bounding boxes for each person are obtained by passing the image through a network. In the reconstruction, the parameters of the SMPL model are estimated for each bounding box $i$ through another network. Then, the mesh of each person is obtained and translated using the vector $t_i$ (Equation \eqref{translation_vector}).

\begin{equation}
    t_i = \begin{bmatrix} 
    \dfrac{d_i (x_i \alpha_i + c_{i,x} - \dfrac{w}{2})}{f} \\\dfrac{d_i (y_i \alpha_i + c_{i,y} - \dfrac{h}{2})}{f} \\ \\ d_i \\ \end{bmatrix},
    \label{translation_vector}
\end{equation}
where $c_{i,x}$ and $c_{i,y}$ are the $x$ and $y$ coordinates of the bounding box centre, $w$ is the width of the image, $h$ is the height of the image, $\alpha_i= max(x_{max} - x_{min}, y_{max} - y_{min})$ is the size of the bounding box, $f$ is the focal length, and $d_i$ is the depth of each person calculated with: 

\begin{equation}
    d_i = \dfrac{\mu f}{s_i \alpha_i},
    \label{depth}
\end{equation}
where  $\mu = 2$ is a multiplicative factor defined by the authors and $s_i$ is an intrinsic parameter predicted by the model. 


Instead of a multi-stage approach, in \cite{romp} one-stage approach is proposed that has a holistic view of the whole image without focusing on each bounding box separately. Therefore, for each image, three distinct maps are acquired: (i) a body centre heatmap, with the location of the 2D body centre locations; (ii) a camera map, and (iii) an SMPL map. These last two maps collectively describe the mesh parameterization. For each location of 2D body centres, their system samples the 3D mesh parameter that serves as input for the SMPL model to generate 3D body meshes. However, this method was exclusively trained with adults. To adapt it to children, \cite{bev} blended the SMPL model with the infant version of it (the SMIL model \cite{hesse2018learning}) through a parameter related to the age group. This parameter was trained with a new dataset which also considered children. Moreover, they introduced a Bird's eye view (BEV) map to reason about the depth. This map represents the likely centres of bodies in depth. 

Our main contributions are the following: (1) the extension of a 3D reconstruction model to be able to deal with children, distinct from the complex end-to-end approaches described before; and (2) the test of a 3D reconstruction model in an unconstrained scenario present in real therapy sessions, which go beyond the data to which this kind of models was trained and is typically applied.

\section{Methods}
In this section we explain how we extended the CRMH model to consider also children. We start by showing the main problems of this method in Section \ref{sec:probCRMH}, then we describe the construction of our regression model to estimate the input $f$ of the CRMH system (Section \ref{sec:fest}). Finally, we present the evaluation protocols and metrics that we used to evaluate our CRMH-p model (Sections \ref{sec:labt} and \ref{sec:realt}.

\subsection{3D reconstruction}
\label{sec:probCRMH}


As a first step to obtain real distances, we started by calculating the focal length. Therefore, we performed one experiment where one adult would stay straight in one known position in front of the camera. Then, we use the model to process this image and calculate the camera intrinsic parameter, $s$, and the bounding box's size, $\alpha$. Finally, the focal length is obtained using \eqref{depth}. 

\begin{figure}[!h]
	\centering
	\includegraphics[width=0.48\textwidth]{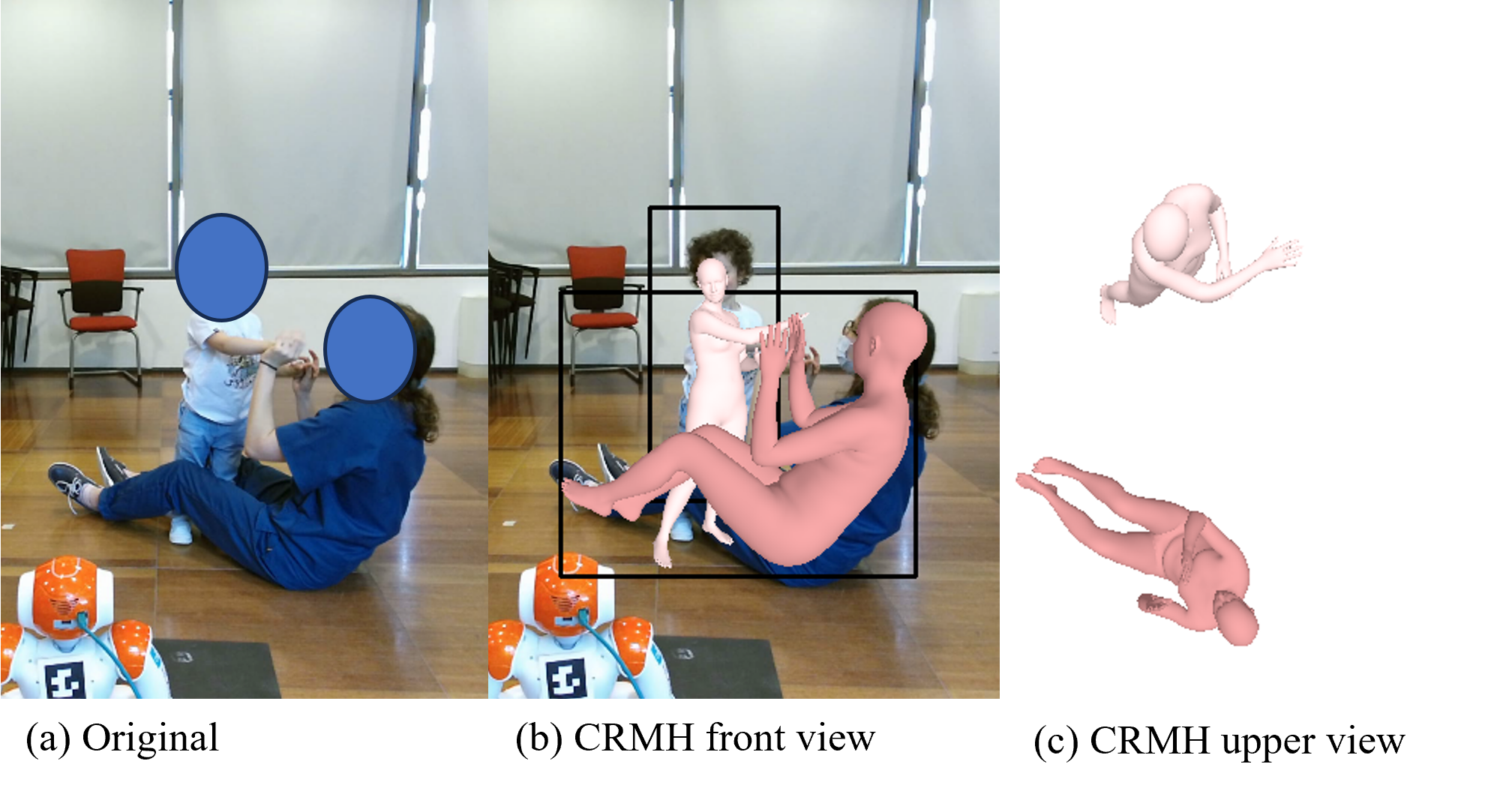}
	\caption{Example of the CRMH problems during therapy sessions with children. Although in the front view (b), the meshes of the child and therapist are correctly identified, in (c) we observe an incorrect translation of child mesh. To account for the smaller dimension of the child, the mesh is reconstructed as an adult with an increased depth.} 
	\label{CRMHp}
\end{figure}

When applying the calculated $f$ to the images of sessions involving children, we observe a translation between the child and the adult who were at the same depth (Figure \ref{CRMHp}). Since the translation vector of \eqref{translation_vector} depends directly on the parameter $f$, we personalise the value $f$ to adapt the model estimates to the children. Empirically we verified that the value of $f$ changed with the height of people. Therefore, we constructed a regression model for estimating $f$ based on the participant's height.

\subsection{Focal-length ($f$) estimation}
\label{sec:fest}

To estimate $f$, we selected a regression model that allows for both offline and online estimation, ensuring rapid execution times. For this model, we used data from therapeutic sessions where a NAO robot, a therapist, and a child with autism played imitation games to train gestures \cite{2021}. The videos were acquired with a Kinect.

To create the regression model, we used the RANSAC method to associate the focal lengths and heights of six therapists who participated in our study. To calculate the focal length for each therapist, we selected a variable window of frames in which they were at 'known' positions and calculated the respective $f$ parameter. The positions chosen were 2.2m  in depth for four of the therapists, 2.5m and 3.1m for the other two. Then, as a loss function to decide the consensus set, we used the weighted sum of squares. The weights considered the uncertainty associated with each of our known positions. This uncertainty was caused by the perspective projection, thus using the parameter $f$ as the focal length, the uncertainty in the depth direction is given by Equation \eqref{eq:uncertainty}, where $h$ is the height of the person and $Z$ is the respective depth.
 \begin{equation}
 \label{eq:uncertainty}
 \begin{split}
     y=\frac{hf}{Z}\Longleftrightarrow
     \frac{\delta y}{\delta Z}=-\frac{hf}{Z^2}\Longleftrightarrow
     {\delta Z} = -\frac{Z^2}{hf}\delta y
 \end{split}   
 \end{equation}

 Therefore, our loss function is given by Equation \eqref{eq:loss function}, in which $w_i=\frac{1}{\sigma_i^2}$, $\sigma_i \propto Z_i^2$ and $f_i$ represents the focal length measured for each therapist and $\hat{f_i}$ the focal length estimated by the regression model.
 \begin{equation}
 \label{eq:loss function}
     L=\frac{\sum_i w_i (f_i-\hat{f}_i)^2}{\sum_i w_i}
 \end{equation}

\subsection{Controlled Experimental Setting}
\label{sec:labt}
To evaluate the accuracy of the regression model, we designed an experiment involving three participants of varying heights: 1.41m (child), 1.62m, and 1.75m. Our objective was to compare the performance of CRMH-personalised (CRMH-p) model with a motion capture system (Optitrack \cite{optitrack}). We decided to also evaluate the more recent BEV model to compare with our CRMH-p model, since it is a model specifically constructed for children. The OptiTrack system served as the ground truth reference for the evaluation of both systems due to its high accuracy and precision. Its sub-millimetre accuracy is given by 12 cameras placed around a well-lit room.


In our experimental protocol, each participant started by standing 1.3m away from the Kinect camera with their arms open. After maintaining this position for 5 seconds, they were instructed to walk backwards while facing the camera until they reached the next designated position. The distances between the designated positions were 1.3m, 1.9m, 2.2m, 2.5m, and 3.0m. Each participant performed the protocol individually to ensure accurate and controlled data collection.

\begin{figure*}

\includegraphics[width=1\textwidth]{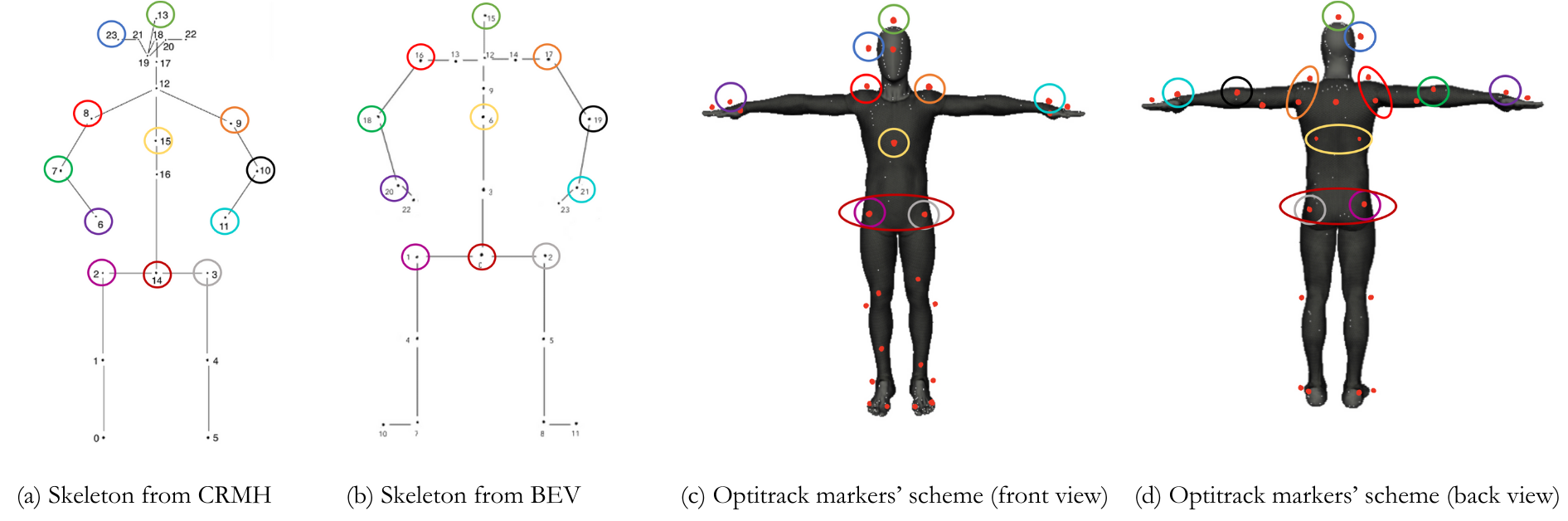}
\caption{ Skeletons’ scheme of the systems. The correspondence between joints used to evaluate are marked by circles of the same colour. Since the systems' skeletons are not a perfect match, for some joints (for example 14 in (a)) we had to use the mean of two associated markers.} 
\label{fig:common-joints}

\end{figure*}

To evaluate the accuracy of the systems, we conducted an analysis based on the coordinates of the skeleton's upper body joints. We used the joint regressor $J$ from SMPL to transform the mesh of CRMH into a 3D skeleton. The BEV model already provided the 3D skeletons. Figure \ref{fig:common-joints} shows the joints used to compare both systems according to their respective 3D skeletons.

The metric used for comparison was the average 3D root mean square error (RMSE) obtained by calculating the RMSE for each frame, as shown in Equation \eqref{eq:rmse}, and then computing the average across all frames. 
\begin{equation}
RMSE = \sqrt{\frac{1}{N}\sum_{i=1}^{N}\left((x_i - \hat{x}_i)^2 + (y_i - \hat{y}_i)^2 + (z_i - \hat{z}_i)^2\right)},
\label{eq:rmse}
\end{equation}
where N represents the total number of joints selected from the skeleton (Figure \ref{fig:common-joints}), ($x_i$, $y_i$, $z_i$) represents the coordinates of each joint position from the ground truth, and ($\hat{x}_i$, $\hat{y}_i$, $\hat{z}_i$) represents the coordinates of each corresponding estimated joint position from the proposed model.

This metric allowed us to assess the differences between the estimated positions obtained from both systems, providing an overall measure of accuracy for the constructed model within a constrained environment. Furthermore, the operational feasibility of each model in real-time scenarios is scrutinised by evaluating their frames per second (FPS) rates. 


We also performed a concise depth analysis focusing on the z-axis (depth direction). We measured the evolution of the child's hip joint's depth throughout the experience to be able to compare the original CRMH model with our proposed model alongside with the ground truth and visualise the impact of adjusting the parameter $f$ individually on the CRMH's depth estimations. This joint was chosen as a representative point, since it had a clear movement in z. 

To ensure the comparability of world coordinates between the CRMH and OptiTrack systems, we estimated a linear model to correct the Kinect camera angle along the y-axis. This adjustment was necessary to align the heights of the skeletons obtained from both systems, enabling accurate and meaningful comparisons. 


\subsection{Real World Setting}
\label{sec:realt}

To test the proposed model, we selected recordings of four therapeutic sessions different from the ones used for the training described in Section \ref{sec:fest} and estimated the focal length of the therapist and child based on their heights. The children chosen were 2 and 3 years old.

For each video, we applied the CRMH model changing the translation vector for the child and the adult using the estimated focal length. We tracked the adult based on a blue suit that all therapists had for working (as shown in Figure \ref{CRMHp} (a)), and the child was the other person in the scene.

To evaluate the constructed model, we compared the number of skeletons identified by the CRMH, BEV and Kinect during the actual session. More precisely, we calculated the percentage of frames from the ground truth where both skeletons (therapist and child) were detected by each of the systems. To establish this ground truth, we used the Segment Anything algorithm \cite{kirillov2023segment} in a selection of frames and extracted the silhouette of the therapist and of the child.  The ground truth was only available in 2D since we were dealing with actual therapy sessions, thus motion capture systems could not be applied.

Therefore, to evaluate the accuracy, we transformed the skeletons obtained with each method, into silhouettes using a dilation operation with a structuring element defined by a circle with the radius of the neck of the ground truth. Finally, using the Dice similarity metric, we compared the silhouette produced by the depth camera and our system with the (2D) ground truth. To consider just the orientation of the skeletons, we have also calculated the centroids of the ground truth and each of the analysed systems and compensated for the offsets between the centroids. 

\section{Experimental Results}

In this section we follow the same structure as the previous section, starting by presenting the linear model used to estimate the focal length (Section \ref{subsec:f_estimation}). Then we compared our CRMH-p model to a state of the art method, in terms of error in relation to the ground truth and processing time in a controlled environment (Section \ref{subsec:control}). The final subsection shows the performance of our method in real world acquisitions. Since in this case there was no ground truth we established our own metric of error and also consider the number of skeletons reconstructed (Section \ref{subsec:real}).

\subsection{The f estimation}
\label{subsec:f_estimation}
Using the RANSAC method and the heights of six therapists, we chose the model with the highest coefficient of determination presented in Table \ref{tab:resultsransac} in green. This table shows also the difference in terms of robustness depending on whether or not the depth-related uncertainty is used in the RANSAC loss function.

\begin{table}[htbp]
\centering
\begin{tabular}{c|r|r|r|}
\cline{2-4}
\multicolumn{1}{l|}{\cellcolor[HTML]{FFFFFF}{\color[HTML]{222222} }}                                      & $R^2$           & \multicolumn{1}{l|}{Slope}     & \multicolumn{1}{l|}{Intercept} \\ \hline
\multicolumn{1}{|c|}{}                                                                                    & 0.7660                         & 250.51                         & -4.51                          \\ \cline{2-4} 
\multicolumn{1}{|c|}{}                                                                                    & 0.7548                         & 271.56                         & -38.58                         \\ \cline{2-4} 
\multicolumn{1}{|c|}{\multirow{-3}{*}{Sum of Squares}}                                                    & 0.9780                         & 696.14                         & -750.66                        \\ \hline
\multicolumn{1}{|c|}{}                                                                                    & 0.9943                         & 158.20                         & 145.72                         \\ \cline{2-4} 
\multicolumn{1}{|c|}{}                                                                                    & \cellcolor[HTML]{B6D7A8}0.9959 & \cellcolor[HTML]{B6D7A8}164.47 & \cellcolor[HTML]{B6D7A8}135.23 \\  \cline{2-4} 
\multicolumn{1}{|c|}{\multirow{-3}{*}{\begin{tabular}[c]{@{}c@{}}Weighted\\ sum of squares\end{tabular}}} & \cellcolor[HTML]{B6D7A8}0.9959 & \cellcolor[HTML]{B6D7A8}164.47 & \cellcolor[HTML]{B6D7A8}135.23                         \\ \hline
\end{tabular}
\caption{Results from the RANSAC model using two different loss functions (sum of squares or weighted sum of squares). Each row corresponds to a linear model ($f=Slope\times height+Intercept$) generated by the RANSAC model with a coefficient of determination $R^2$. The selected model is underlined in green.}
\label{tab:resultsransac}
\end{table}

\subsection{Controlled Experimental Setting}
\label{subsec:control}
We present the results obtained from the controlled testing phase, to evaluate the performance and accuracy of the regression model for 3D skeleton prediction. Table \ref{tab:focal-height} provides an overview of the height and respective focal lengths of the participants in the study according to the chosen model in Section \ref{subsec:f_estimation}.

\begin{table}[htbp]

\centering
\begin{tabular}{|c|c|}
\hline
Height ($m$) & Focal Length \\
\hline
1.41 (Child) & 367.13 \\
\hline
1.62 & 401.67 \\
\hline
1.75 & 423.05 \\
\hline
\end{tabular}
\caption{Height and Focal Lengths of participants.}
\label{tab:focal-height}
\end{table}

With those focal length values, we compare the results of the CRMH-p model with the BEV model in terms of accuracy and computational load. 

\begin{table}[htbp]
\centering
\begin{tabular}{c|c|c|c|c|}
\cline{2-5}
& \multicolumn{2}{c|}{RMSE ($m$)}& \multicolumn{2}{c|}{FPS }\\
\hline
\multicolumn{1}{|c|}{Height ($m$)} & CRMH-p&BEV& CRMH-p&BEV\\
\hline
\multicolumn{1}{|c|}{1.41} & \textbf{0.20}&0.35& \textbf{8.02}&5.25\\
\hline
\multicolumn{1}{|c|}{1.62} & \textbf{0.23}&0.27& \textbf{7.96}&5.24\\
\hline
\multicolumn{1}{|c|}{1.75} & \textbf{0.14} &0.19& \textbf{7.90}&5.20\\
\hline
\end{tabular}

\caption{3D RMSE results and processing rate for the CRMH-p and BEV model for each participant in the depth $1.9m<z<2.5m$ corresponding with the region where therapies take place. The best results for each height are marked in bold.}
\label{tab:rmse-results}
\end{table}

Table \ref{tab:rmse-results} shows that both models perform better for the tallest person, both in terms of accuracy and processing rate. Furthermore, for the three people, the CRMH-p model is more accurate than BEV in the depth-range that is used in typical therapy sessions ($1.9m<z<2.5m$). Moreover, the CRMH-p model is faster than the BEV model.


Overall, our method has an error below one body width (0.3m), which is considered acceptable for this kind of application \cite{bev}. However, part of the observed error in Table \ref{tab:rmse-results} is due to the dissimilarity in joint representations among the 3D skeletons obtained from each system, see Figure \ref{fig:common-joints}.

\begin{figure}[!h]
	\centering
	\includegraphics[width=0.35\textwidth]{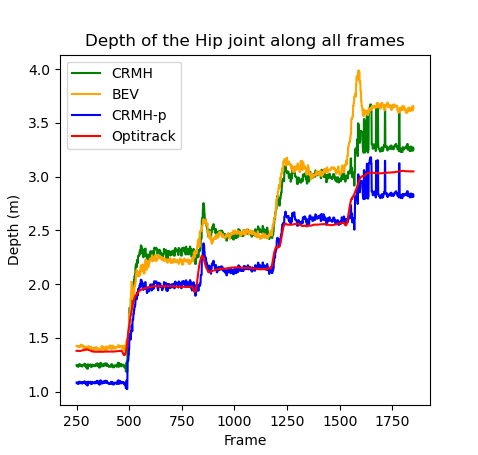}
	\caption{Depth values for the child's hip joint using different systems: CRMH,  BEV, CRMH-personalised (CRMH-p) and Optitrack. The proposed model improved the performance of the original CRMH. In the middle depth range ($1.9m<z<2.5m$) the accuracy of the proposed model is notably high. } 
	\label{fig:depth-analysis}
\end{figure}

Following, to further assess the impact of the proposed model, we analyzed the depth estimates of the child's hip (Figure \ref{fig:depth-analysis}). The analysis involved four different systems: the OptiTrack system (serving as the ground truth reference), the CRMH-p model using participant-specific focal lengths, the original CRMH model using the focal length proposed by the authors, and the BEV model.

Figure \ref{fig:depth-analysis} shows the benefits of using the regression model with personalised focal lengths according to each person's height. 
The figure illustrates the improved accuracy achieved by the proposed model in estimating joint positions. Our model is more accurate than BEV in all depth ranges except for the smaller depth ranges. Moreover, within the middle depth range, the depth more commonly used in therapies, the accuracy of the proposed model is higher, further validating its performance.






\subsection{Real World Setting}
\label{subsec:real}

The results of the focal lengths for the four children analyzed are shown in Table \ref{tab:focalca} together with the focal lengths of the respective therapists (adults). To compensate for their smaller height, the focal lengths of the children are considerably smaller than the adults. 

\begin{table}[htbp]
\begin{tabular}{ll|r|ll|r|}
\cline{3-3} \cline{6-6}
                                                &    & \multicolumn{1}{l|}{Focal length} &                                                  &     & \multicolumn{1}{l|}{Focal length} \\ \hline
\multicolumn{1}{|l|}{\multirow{4}{*}{Children}} & 17 & 282.45                            & \multicolumn{1}{l|}{\multirow{4}{*}{Therapists}} & PJ  & 414.83                            \\ \cline{2-3} \cline{5-6} 
\multicolumn{1}{|l|}{}                          & 08 & 290.21                            & \multicolumn{1}{l|}{}                            & ARG & 411.54                            \\ \cline{2-3} \cline{5-6} 
\multicolumn{1}{|l|}{}                          & 02 & 274.68                            & \multicolumn{1}{l|}{}                            & SF  & 398.38                            \\ \cline{2-3} \cline{5-6} 
\multicolumn{1}{|l|}{}                          & 11 & 309.62                            & \multicolumn{1}{l|}{}                            & MJC & 403.32                            \\ \hline
\end{tabular}
\caption{Focal lengths of the children and adults analyzed. The names of the children and therapists are represented by a code for anonymization purposes.}
\label{tab:focalca}
\end{table}

\begin{table*}[htbp]
\centering
\begin{tabular}{l|rrr|rrr|}
\cline{2-7}
                                   & \multicolumn{3}{c|}{Therapist}                                 & \multicolumn{3}{c|}{Child}                                     \\ \cline{2-7} 
                                   & \multicolumn{1}{c|}{Kinect} & \multicolumn{1}{c|}{CRMH-p}& \multicolumn{1}{c|}{BEV} & \multicolumn{1}{c|}{Kinect} & \multicolumn{1}{c|}{CRMH-p}& \multicolumn{1}{c|}{BEV} \\ \hline
\multicolumn{1}{|l|}{MJC 17} & \multicolumn{1}{r|}{0.092}  &  \multicolumn{1}{r|}{\textbf{0.228}} & 0.189                            & \multicolumn{1}{r|}{0.150}  &  \multicolumn{1}{r|}{\textbf{0.165}} &0.128                            \\ \hline
\multicolumn{1}{|l|}{PJ 08}  & \multicolumn{1}{r|}{0.232}  &  \multicolumn{1}{r|}{\textbf{0.242}} &0.235                           & \multicolumn{1}{r|}{\textbf{0.227}}  &  \multicolumn{1}{r|}{0.216}&0.198                            \\ \hline
\multicolumn{1}{|l|}{SF 02}  & \multicolumn{1}{r|}{0.225}  &  \multicolumn{1}{r|}{\textbf{
0.239}}&0.223                            & \multicolumn{1}{r|}{0.227}  &  \multicolumn{1}{r|}{\textbf{0.234}} &0.192                           \\ \hline
\multicolumn{1}{|l|}{ARG 11} & \multicolumn{1}{r|}{0.241}  &  \multicolumn{1}{r|}{\textbf{0.245}} &0.223                           & \multicolumn{1}{r|}{\textbf{0.228}}  &  \multicolumn{1}{r|}{0.227} &0.201                           \\ \hline
\end{tabular}
\caption{Mean dice similarity between the skeletons of the ground truth and the ones of the analysed system (Kinect or CRMH-personalised) with the centroid correction. Each row represents one session done by a therapist (characters code) and a child (numerical code). The best results for each child and for each therapist are marked in bold.}
\label{tab:twodicecent}
\end{table*}
From the percentage of detected two skeletons (Table \ref{tab:twoskeletons}), the proposed model and the BEV model were able to reconstruct several of the skeletons lost by Kinect. 
\begin{table}[h!]
\centering
\begin{tabular}{|l|r|r|r|}
\hline
Two skeletons & \multicolumn{1}{l|}{\% Kinect} & \multicolumn{1}{l|}{\% CRMH-p} & \multicolumn{1}{l|}{\% BEV} \\ \hline
MJC 17  & 12                             & 92& \textbf{99}                                \\ \hline
PJ 08   & 10                             & 75& \textbf{87}                                  \\ \hline
SF 02   & 57                             & 77 & \textbf{99}                                 \\ \hline
ARG 11  & 83                             & \textbf{94}  & 73                                \\ \hline
\end{tabular}
\caption{Percentage of ground truth skeletons in which both skeletons were detected by the Kinect and the proposed model. Each row represents one session done by a therapist (characters code) and a child (numerical code). The best results for each pair (therapist-child) are marked in bold.}
\label{tab:twoskeletons}
\end{table}
Regarding the Dice similarity in the skeletons identified by both algorithms (Table \ref{tab:twodicecent}) and considering just the orientation, our system had similar results to the Kinect, outperforming it for most of the cases. BEV had the worse performance of the three compared systems.

We now analyze the skeletons reconstructed by our system and that could not be identified by the Kinect. The values of dice similarities were similar to the ones of skeletons identified by both algorithms (Table \ref{tab:onedice}), showing that the algorithm generalizes well to the more challenging cases where the Kinect fails. 
\begin{table}[h!]
\centering
\begin{tabular}{l|rr|}
\cline{2-3}
                                   & \multicolumn{2}{c|}{CRMH-p}                            \\ \cline{2-3} 
                                   & \multicolumn{1}{l|}{Therapist} & \multicolumn{1}{l|}{Child} \\ \hline
\multicolumn{1}{|l|}{MJC 17} & \multicolumn{1}{r|}{0.245}     & 0.149                      \\ \hline
\multicolumn{1}{|l|}{PJ 08}  & \multicolumn{1}{r|}{0.236}     & 0.188                      \\ \hline
\multicolumn{1}{|l|}{SF 02}  & \multicolumn{1}{r|}{0.231}     & 0.202                      \\ \hline
\multicolumn{1}{|l|}{ARG 11} & \multicolumn{1}{r|}{0.231}     & 0.181                      \\ \hline
\end{tabular}
\caption{Mean Dice similarity for the skeletons that were identified by the proposed model but not by Kinect. Each row represents one session done by a therapist (characters code) and a child (numerical code).}
\label{tab:onedice}
\end{table}

\section{Conclusions and Future Work}

We propose an extension to a state-of-the-art method (CRMH ) for 3D pose reconstruction to be able to deal with images with children. This model is simpler than the end-to-end approaches available in the literature, and it is capable of correctly repositioning children less than five years old, age groups absent from the training set of the original model. 

In controlled settings, the proposed model 3D error is always lower than $0.3m$ in 3D, when compared with the Optitrack system, the goal standard in biomechanics evaluations and lower than the recent BEV model. Moreover, it has a lower processing time, being preferable for a future online application.
The depth error is low for the normal operation distances and remains almost constant outside this range, therefore it can be used in highly unconstrained scenarios as during therapy sessions.
Indeed, in real-world settings, we demonstrate that the proposed model performs comparably to a depth camera (Kinect) and excels in recovering a significant portion of the frames lost by Kinect.

Our future work will focus on the optimization of the geometry of the acquisition setup to further improve accuracy, as well as extend the model to the online setting.

\bibliography{egbib}

\end{document}